\pdfoutput=1

\documentclass[11pt]{article}
\usepackage[table]{xcolor}
\usepackage[]{acl}
\usepackage{pifont}
\usepackage{subcaption}

\usepackage{times}
\usepackage{latexsym}
\usepackage{adjustbox}
\usepackage{pdfpages}
\usepackage{arydshln}

\usepackage{amsmath}
\usepackage[T1]{fontenc}
\usepackage{tikz}
\usetikzlibrary{fit,tikzmark}
 
\usepackage[utf8]{inputenc}

\usepackage{microtype}

\title{Development of Cognitive Intelligence in Pre-trained Language Models}

\makeatletter
\def\thanks#1{\protected@xdef\@thanks{\@thanks
        \protect\footnotetext{#1}}}
\makeatother

\newcommand{\gtlogo}{\raisebox{3.4pt}{\includegraphics[scale=0.04]{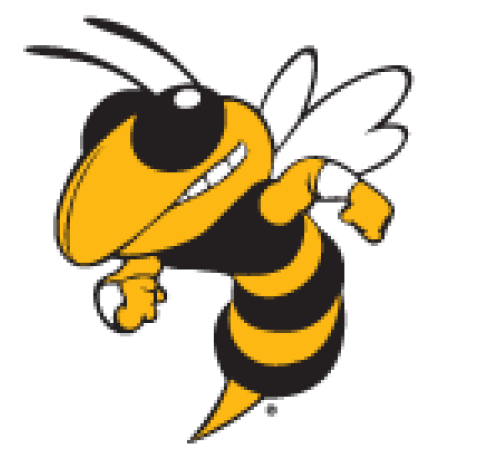}}}
\newcommand\tc[1]{\textcolor{blue}{#1}}

\author{Raj Sanjay Shah, Khushi Bhardwaj, Sashank Varma
 \\ 
 Georgia Institute of Technology \gtlogo \\
 \textcolor{darkblue}{{\{\href{mailto:rajsanjayshah@gatech.edu}{rajsanjayshah},
\href{mailto:khushi.bhardwaj@gatech.edu}{khushi.bhardwaj}, \href{mailto:varma@gatech.edu}{varma}\}@gatech.edu}
}}
\begin{document}
\maketitle
\begin{abstract}





Recent studies show evidence for emergent cognitive abilities in Large Pre-trained Language Models (PLMs). The increasing \emph{cognitive alignment} of these models has made them candidates for cognitive science theories. Prior research into the emergent cognitive abilities of PLMs has largely been \emph{path independent} to model training, i.e., has focused on the final model weights and not the intermediate steps. However, building plausible models of human cognition using PLMs would benefit from considering the \emph{developmental alignment} of their performance during training to the trajectories of children's thinking. 
Guided by psychometric tests of human intelligence, we choose four sets of tasks to investigate the alignment of ten popular families of PLMs and evaluate their available \emph{intermediate and final training steps}. These tasks are Numerical ability, Linguistic abilities, Conceptual understanding, and Fluid reasoning. We find a striking regularity: regardless of model size, the developmental trajectories of PLMs consistently exhibit a window of maximal alignment to human cognitive development. 
Before that window, training appears to endow ``blank slate'' models with the requisite structure to be poised to rapidly learn from experience. After that window, training appears to serve the engineering goal of reducing loss but not the scientific goal of increasing alignment with human cognition.

\end{abstract}

\section{Introduction}
\begin{figure*}[htp]
\centering
\includegraphics[trim={2cm 2cm 2cm 2cm}, width=0.95\textwidth]{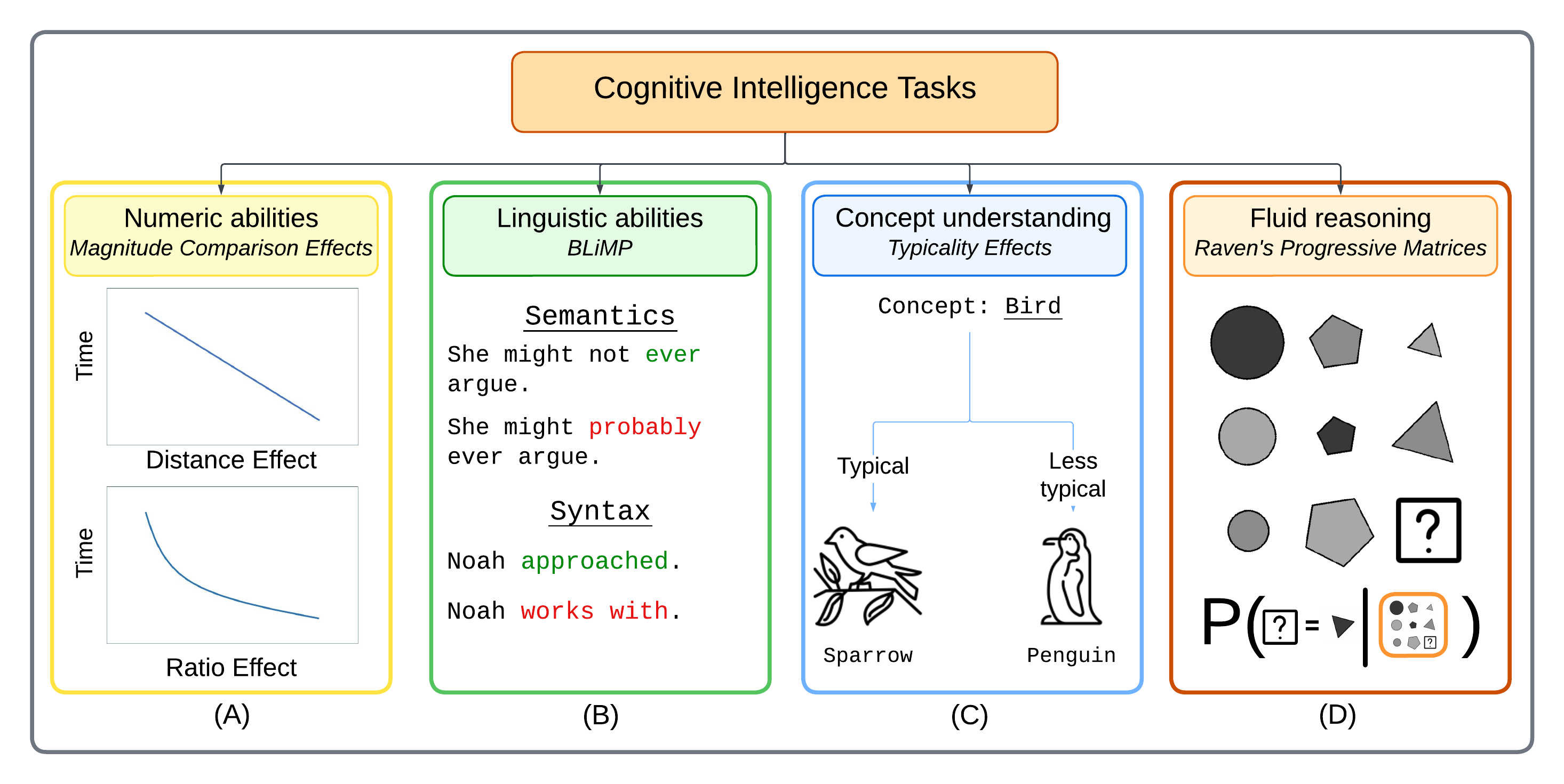}
\caption{A list of cognitive intelligence tasks under consideration.}
\label{fig:introduction0}
\end{figure*}

Large Pre-trained Language Models (PLMs) like Google's Gemini \cite{geminiteam2023gemini}, Meta's LLaMA 2 \cite{touvron2023llama}, and OpenAI's GPT 4 \cite{openai2023gpt4} show human-level or even super-human performance on many cognitive performance tasks.
This is true in domains such as mathematical reasoning \cite{shah-etal-2023-numeric, ahn2024large}, language comprehension \cite{warstadt2020blimp, ye2023comprehensive, koubaa2023gpt}, concept understanding \cite{cog_sci_typicallity}, and analogical reasoning \cite{webb2023emergent, hu2023context}. These successes have contributed to the hype of reaching Artificial General Intelligence (AGI). 



Such claims deserve to be scrutinized.
There is a massive disparity between the training data scale of PLMs and humans. However, PLMs unintentionally acquire human performance characteristics from the corpora they are trained on, through residues of the values, beliefs, and biases of the authors of the texts \cite{ai_psychometry}.

We approach the human alignment of PLMs by grounding their evaluation in frameworks for \emph{psychometric intelligence}. Psychometric measures of intelligence include multiple subtests spanning a range of abilities, including mathematical thinking, language comprehension, spatial thinking, and fluid reasoning
\cite{snow1984topography, 
 carroll1993human, sternberg2000handbook, mcgrew2009chc, haier2023neuroscience}. 
In this work, we choose representative assessments of different facets of human intelligence, modified for the required textual modality, to evaluate the \emph{cognitive alignment} of PLMs. 

A second goal of our work is to move beyond cognitive alignment to also evaluate the \emph{developmental alignment} of PLMs.
The claim that the final model state of a PLM approximates adult performance leaves open the question of the path by which it arrived there. Ideally, the model's performance improvements over training should also track the progression of cognitive abilities over development \cite{elman1996rethinking, bengio2009curriculum}. This potential parallelism would be stronger evidence for PLMs as cognitive science models. Researchers are increasingly addressing this question by building PLMs trained on a developmentally plausible corpus of child-directed speech, transcribed dialogue, and children’s literature \cite{huebner-etal-2021-babyberta, conll-2023-babylm, bhardwaj2024pretraining}. 

We ask the question of developmental alignment in a theoretically important way: Is the cognitive alignment of PLMs achieved in a \emph{path-independent} or \emph{path-dependent} manner? Prior studies focusing on the cognitive alignment of PLMs have only established path independence: that models at the end of training approximate adult performance across a range of domains. Here, we also evaluate path dependence: Do the performance improvements of PLMs over training track the growth of these abilities in children over development?
We ask this question for models of different sizes and track their developmental alignment over millions and billions of training tokens.


\noindent To summarize, our key contributions are as follows:
\begin{itemize}
    \setlength\itemsep{0em}
    \setlength\parskip{0em}
    \setlength\parsep{-2em}
    \item \textbf{Cognitive Modelling using AI:} We test the appropriateness of PLMs for cognitive modeling by evaluating whether their performance profiles match those of adults.
    \item \textbf{Developmental trajectories in LLM pre-training and scaling:} Previous studies have largely focused on evaluating the final training checkpoints of PLMs for their cognitive plausibility, and have neglected the question of developmental trajectories. Here, we also ask: Can PLMs be used to model developmental trajectories of children's thinking despite the training data scale mismatch?
    \item \textbf{Representative tasks:} We choose representative tasks of human cognition taken from psychometric tests of intelligence tests. These tasks evaluate numeric, linguistic, conceptual, and fluid intelligence. We propose these to be a \emph{prerequisite} for considering PLMs as cognitive science models.
\end{itemize}

\begin{table*}[!h]
\centering

\resizebox{0.9\textwidth}{!}{%
\begin{tabular}{llll}
\hline
Cognitive Domain & Task            & Source                    & License    \\

\hline
Numeric Abilities & Magnitude Comparison Effects & \cite{shah-etal-2023-numeric} &    (cc by 4.0)  \\
Linguistic Abilities & BLiMP & \cite{warstadt2020blimp}  & (cc by 4.0) \\
Concept Understanding & Typicality Effects & \cite{cog_sci_typicallity, castro21}  & (cc by 4.0) \\
Fluid reasoning & Raven's Progressive Matrices & \cite{hu2023context} &  (cc by 4.0) \\
\hline
\end{tabular}
}
\caption{Summary of assessments.}

\label{tab:ThinkTank_tasks}
\vspace*{-1em}
\end{table*}

\section{Related work}

\subsection{Psychometric theories of intelligence}


Previous studies of the human alignment of ML models have typically looked at singular dimensions, such as numeric abilities \cite{zhuang2023efficiently, fang2024patch}.  
Rather than choose cognitive abilities in a piecemeal fashion, we look to psychometric theories of intelligence for guidance \cite{sternberg2000handbook}. These theories distill performance on a large number of subtests into a small number of latent factors.
Despite popular attention to ``general intelligence'' and the latent factor \emph{g}, there is a long history of theories positing that intelligence is composed of multiple domain-specific abilities. An important, early domain-specific theory of intelligence included seven ``primary abilities'' \cite{thurstone1938primary}. The most widespread psychometric theory of intelligence today, the Cattell-Horn-Carrol (CHC) theory \cite{carroll1993human,mcgrew2009chc}, includes among its ``broad'' abilities quantitative knowledge, reading and writing ability, fluid reasoning, and ``comprehension'' knowledge (a subcomponent of which is conceptual understanding). We evaluate the cognitive and developmental alignment of PLMs along these four abilities.

\subsection{Emergent cognitive abilities in Language Models}

Recently, the performance of language models has improved as they have increased in size from millions to billions of parameters, trained on larger corpora, and further tuned in novel ways (instruction tuned, RLHF). This has led researchers to increasingly advocate for the use of PLMs as cognitive models \cite{piantadosi2023modern,warstadt2024artificial}. Increasing the number of parameters of the models has given rise to \emph{emergent abilities} that cannot be predicted by extrapolating from the performance of smaller models \cite{wei2022emergent}. Emergent abilities have been observed in a variety of task types such as multi-task language understanding \cite{hendrycks2021measuring}, grounded conceptual mapping \cite{patel2022mapping}, and truthfulness \cite{lin2021truthfulqa}.
In recent work, \citet{hoffmann2022training} and \citet{biderman2023pythia} have shown the benefits of training a model for more tokens on problem-solving \cite{wei2022chain}, common-sense reasoning \cite{sakaguchi2021winogrande}, arithmetic abilities \cite{biderman2023pythia}, and linguistic performance \cite{paperno2016lambada}. Although the presence of emergent abilities extends to cognitive science domains \cite{wei2022chain, goertzel2023generative, hagendorff2023machine}, prior studies have been piecemeal in their approach and have failed to (1) consider multiple cognitive abilities as specified by theories of psychometric intelligence and (2) move beyond cognitive alignment to also evaluate the developmental alignment of PLMs over training.

\subsection{Pre-trained language model use in developmental modeling}
Recently, researchers have begun advocating for the use of PLMs for modeling cognitive development in children \cite{kosoy2023comparing,salewski2024context}. For example, \citet{Portelance2023PredictingAO} and \citet{bhardwaj2024pretraining} suggest the use of language models to predict the age of acquisition of words in children. Researchers have also proposed studying second language acquisition and bilingualism by mapping pre-training steps in PLMs to understand the rate of language development \cite{evanson2023language, marian2023studying, sharma1monolingual}. 
We evaluate the proposal that the performance of intermediate training checkpoints of PLMs maps to points during children's cognitive development.

\section{A suite of psychometric intelligence tasks}

We assemble a suite of tasks that benchmark PLMs across four facets of psychometric intelligence. Table \ref{tab:ThinkTank_tasks} summarizes the tasks along with the licensing details for public use. The details of each assessment and their respective operationalization are given below. \footnote{We add the tasks to a publically available unified language model testing framework, titled \textit{lm-evaluation-harness} \cite{eval-harness}, to support the evaluation of future models on these psychometric intelligence assessments.}

\subsection{Numeric abilities}
\begin{figure}[htp]
\centering
\includegraphics[trim={1cm 2 1 2}, width=0.3\textwidth]{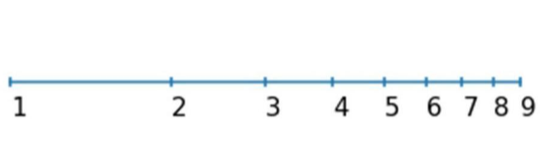}
\caption{ Mental Number Line: Organization of magnitude representations in a logarithmically scaled manner.}
\label{fig:mnl}
\end{figure}

The question of how humans understand symbolic numbers has been investigated by cognitive scientists for more than half a century. These studies show that people map number symbols to a \emph{mental number line} (MNL, Figure \ref{fig:mnl}) with a log-compressed psychophysical scale 
\cite{MOYER1967}.

Prior research on the numerical abilities of PLMs has focused on higher-order and application-driven aspects of mathematics such arithmetic equations and word problems~\cite{ awp_1, awp_2, yuan2023large}, exact facts ~\cite{exact_facts}, and measurement estimation ~\cite{approx_measurement}. \emph{However, these tasks fail to directly tap the foundational cognitive mechanisms underlying human numerical understanding: recruitment of the MNL.}

In a recent study, \citet{shah-etal-2023-numeric} found evidence for a human-like MNL in various PLMs. They showed that despite lacking explicit neural circuitry to represent numbers, through experience (i.e., vast amounts of training data), PLMs show human-like performance profiles and learn latent representations consistent with the MNL.


We follow \citet{shah-etal-2023-numeric} and look for the two behavioral signatures of a compressed MNL representation, the distance effect and the ratio effect. In humans, these are defined as:
\begin{itemize}
    \setlength\itemsep{0em}
    \setlength\parskip{0em}
    \setlength\parsep{0em}
    \item \textbf{Distance effect} (refer to Figure \ref{fig:introduction0}A - top): The greater the distance $|x-y|$ between two numbers $x$ and $y$, the faster they are compared, i.e., the greater (or lesser) number is identified 
    \cite{MOYER1967}.
    \item \textbf{Ratio effect} (refer to Figure \ref{fig:introduction0}A - bottom): The time to compare two numbers $x$ and $y$ decreases nonlinearly as a function of the ratio of the larger number over the smaller number $\frac{max(x,y)}{min(x,y)}$ \cite{halberdaIndividualDifferencesNonverbal2008}.
\end{itemize}

These effects can be mapped to language models by adopting the following linking hypothesis: \emph{the greater the cosine similarity of two number representations in a PLM, the more difficult it is to discriminate them (i.e., to judge which one is greater (or lesser)), and thus the longer it takes}. While we focus on the Distance and Ratio effects in PLMs, the results for all the effects investigated by \citet{shah-etal-2023-numeric} are in Appendix \ref{nums_appendix}.
\newline

\textbf{Operationalization:} We used the same protocol as \citet{shah-etal-2023-numeric}. For each effect, we test three formats of number representations of PLMs: mixed-case number words, lower-case number words, and digits. We present the $R^2$ values for the Distance and Ratio effects, which are averaged across each input representation. The $R^2$ values for the distance effect in PLMs are obtained by fitting a linear function predicting the cosine similarity of $x$ and $y$ from their distance $|x-y|$. $R^2$ values for the ratio effect in PLMs are obtained by fitting a negative exponential function predicting the normalized cosine similarity of $x$ and $y$ from their ratio $\frac{\max(x, y)}{\min(x, y)}$. 
Note: This task requires access to the latent representations of models.




\subsection{Linguistic abilities}







Language (or verbal) ability is a central component of human cognition and cognitive neuroscience \cite{hagoort2019human}. At the dawn of the cognitive revolution, it was conceptualized as a largely innate ability, and language acquisition was thought to require relatively little learning from experience \cite{fodor1985precis,chomsky2014aspects}. More recently, cognitive developmentalists have shown that infants can learn language through exposure to the statistical regularities of the linguistic environment \cite{saffran1996statistical, siegelman2020statistical}. These findings have been modeled using multi-layer perceptrons \cite{elman1996rethinking} and, more recently, PLMs \cite{lake2023word}.

We use BLiMP (Benchmark of Linguistic Minimal Pairs for English) \cite{warstadt2020blimp} to evaluate the linguistic abilities of each PLM under consideration. BLiMP consists of 67 datasets of 1000 pairs of minimally different sentences which vary in acceptability and span 12 phenomena at three levels of language: \textit{morphology}, \textit{syntax}, and \textit{semantics}. The 12 phenomena are described in Appendix \ref{sec:ling_app}. Each pair consists of one acceptable sentence and one unacceptable sentence which otherwise differ minimally. BLiMP evaluates the models by measuring if they assign a higher probability to the acceptable vs. unacceptable sentence of each pair. Figure \ref{fig:introduction0}B shows two examples of minimal pairs. 

\textbf{Operationalization:} We use the LM-eval-harness \cite{eval-harness} benchmarking suite to test our models on the BLiMP tasks. We evaluate if a model assigns a higher sequential probability to the acceptable sentence. Note: This requires models that can generate probabilities of tokens. 

\subsection{Concept understanding}
On encountering a new stimulus, humans categorize it -- assign it to a known concept -- in order to make inferences about its unobservable properties \cite{murphy2002big}. A striking finding is that not all members of a category are equal \cite{rosch75}. Rather, for a given category (e.g., Bird), some members (e.g., pigeon) are more typical than others (e.g., ostrich). This phenomenon, known as the \emph{typicality effect}, is a central feature of human categorization \cite{lakoff2008women}.

Typicality gradients in human categories can be measured using the production task, where participants are given a category label (e.g., Bird) and asked to list as many members of the category as they can in a limited time \cite{battig1969category, van2004category, castro21}. The typicality of an item is defined as the proportion of participants who produce it. 

Language models have shown some evidence of human-like typicality gradients. \citet{heyman2019prediction} used word2vec embeddings to predict the category typicality norms released by \citet{de2008exemplar}. More recent work by \citet{misra2021language} and \citet{bhatia2022transformer} has looked at correlations of PLMs like BERT, RoBERTa, and GPT-2 to the \citet{rosch75} typicality norms for ten categories. \citet{cog_sci_typicallity} performed the most comprehensive study of the alignment of concept understanding in the latent representations of PLMs. We expand upon their task setup to evaluate human-like concept understanding in the PLMs that are the focus here.

 \textbf{Operationalization:} For each model, we calculate the representativeness of a member to its category in three possible ways:
 \begin{itemize}
     \setlength\itemsep{0em}
    \setlength\parskip{0em}
    \setlength\parsep{0em}
     \item Closeness judgment problem: Calculate the cosine similarity between the obtained latent representations for the member and the category. This requires models where the latent representations are readily available.
     \item Surprisal values: For each member of a category, the probability of the sequence \textit{a "member" (eg. pigeon) is a "category" (eg. bird)}. This method requires access to the probability of each token in a sequence.
     \item Prompting:  Prompt the models with the following design: Guidelines, Query, and Options. The Guideline highlights the task of re-ranking the members given in the Options based on appropriateness with the Query. The Query consists of the in-filling task: \textit{A \rule{0.5cm}{0.15mm} is a [category name]}. The Options are each of the possible members of the category. Given the complexity of the prompting, usable outputs are only obtained from models that are larger than 30 billion parameters. 
 \end{itemize}
 
For the two in-filling problems (i.e., based on surprisal values and prompting), we also evaluate models on zero to three exemplars as context. The details of the experiments on these different exemplar contexts are given in Appendix \ref{sec:typ}. 


\subsection{Fluid reasoning}

Humans can logically parse information and detect patterns in novel stimuli without having to rely on prior experiences or learned information. This ability is called fluid reasoning \cite{cattell1963theory}.



\newcommand{\cmark}{\textcolor{green!80!black}{\ding{51}}}
\newcommand{\xmark}{\textcolor{red}{\ding{55}}}

\begin{table*}[!h]
    \centering
    
    \resizebox{\textwidth}{!}{%
    \begin{tabular}{lccccccc}
    \hline
        Models & Source & Latent rep.  & Token prob.  & Multiple sizes   & Intermediate checkpoints   & Known training order \\ 
        \hline
        Amber & \cite{liu2023llm360} &  \cmark &  \cmark &  \xmark & \cmark &  \cmark  \\ 
       
        Falcon & \cite{falcon40b} &  \cmark &  \cmark & \xmark & \xmark & \xmark \\ 
         Starling & \cite{starling2023} &  \cmark &  \cmark & \xmark & \xmark & \xmark \\ 
        Llama & \cite{touvron2023llama} &  \cmark &  \cmark & \cmark & \xmark & \xmark  \\ 
        Mistral & \cite{jiang2023mistral} &  \cmark &  \cmark & \xmark & \xmark & \xmark  \\ 
        Qwen & \cite{qwen} &  \cmark &  \cmark & \cmark & \xmark & \xmark \\ 
        \hline
        \multicolumn{1}{|l}{\begin{tabular}[c]{@{}c@{}}  Pythia  \end{tabular}} & \cite{biderman2023pythia} &  \cmark &  \cmark &  \cmark &  \cmark &   \multicolumn{1}{c|}{\begin{tabular}[c]{@{}c@{}} \cmark  \end{tabular}} \\ 
        \hline
        Gemini & \cite{geminiteam2023gemini} & \xmark & \xmark & \xmark & \xmark & \xmark  \\ 
        GPT-3.5-Turbo & \cite{openai} & \xmark &  \cmark & \xmark & \xmark & \xmark  \\ 
        GPT 4 & \cite{openai2023gpt4} & \xmark &  \cmark & \xmark & \xmark & \xmark \\ 
        \hline
    \end{tabular}
    }
    \caption{List of language model families under consideration with their statistics.}
    \label{tab:models}
\end{table*}

\begin{figure}[h]
\centering
\includegraphics[trim={2cm 1 1 1}, width=0.43\textwidth]{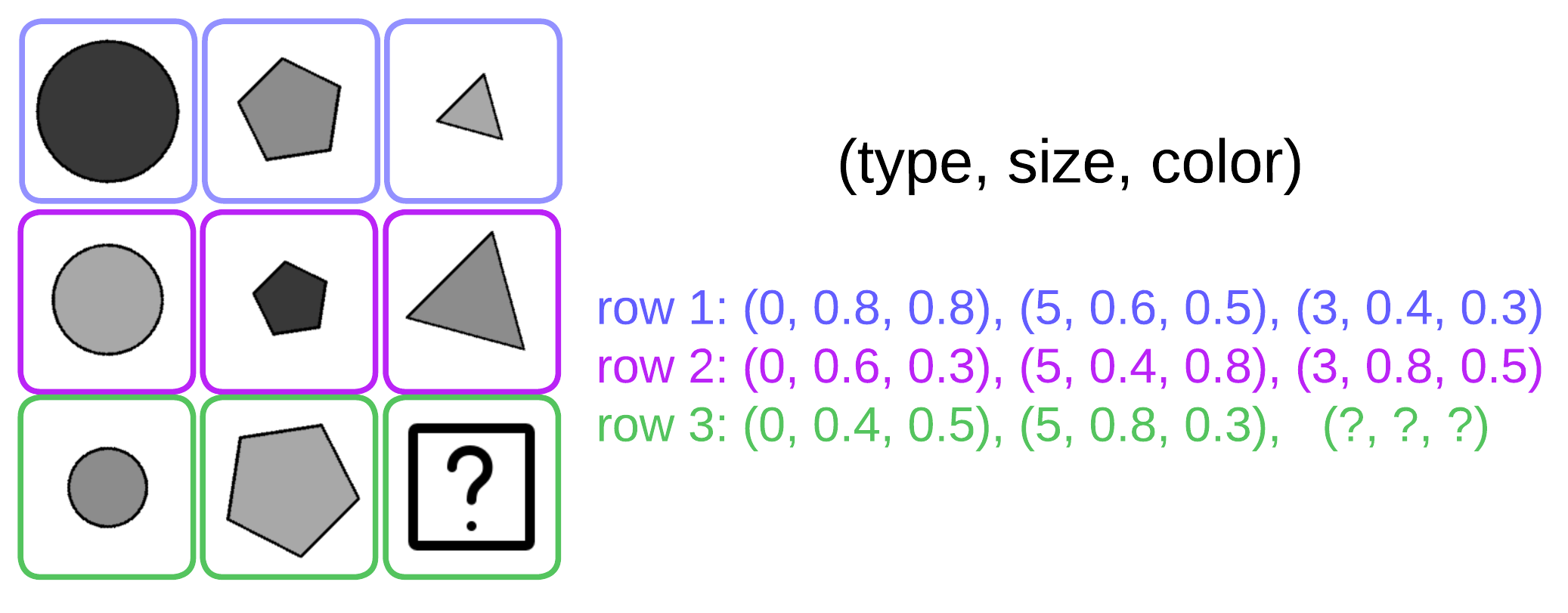}
\caption{Example adaptation of visual RPM problems to the textual format. Each image is decomposed into tuples of (type, size, color). Type indicates the shape of the image.}
\label{fig:conversion_rpm}
\end{figure}

We focus on the dominant measure of fluid reasoning, the Ravens Progressive Matrices (RPM) test \cite{raven2003raven}. An example Ravens-like problem is given in Figures \ref{fig:introduction0}D and \ref{fig:conversion_rpm}. An RPM item consists of a 3x3 matrix of cells with one empty cell. Participants must induce the underlying, abstract patterns that hold across the rows and columns of the matrix, and apply these to infer the image in the empty cell from a given set of options. These images vary in visual attributes like shape and color, along with more abstract qualities. The RPM is the standard measure of fluid reasoning \cite{snow1984topography} and is highly correlated with analogical reason \cite{GOSWAMI198673,webb2023emergent}.

Given the visual nature of the RPM, previous work by \citet{hu2021stratified, hu2023context} and \citet{webb2023emergent} mapped the Raven-10000 dataset to a textual format to facilitate the testing of PLMs. The mapping involves reformulating visual elements into text-based numerical tuples representing attributes like shape, size, and color textually, as illustrated in Figure \ref{fig:conversion_rpm}, to form the I-Raven dataset. We use their approach with a focus on the ``Center Single Alignment'' sub-task, which features a single shape per matrix cell. We differ from their work by evaluating a broader set of models.

\begin{table*}[!h]
    \centering
   
    \resizebox{0.98\textwidth}{!}{%
    \begin{tabular}{l:cc:c:cc:c}
    \hline
        &  \multicolumn{2}{c:}{\begin{tabular}[c]{@{}c@{}}  Numeric Abilities  \end{tabular}} & Linguistic Abilities &  \multicolumn{2}{c:}{\begin{tabular}[c]{@{}c@{}}  Conceptual Understanding  \end{tabular}} & Fluid reasoning \\ \hdashline
        \multicolumn{1}{c:}{\begin{tabular}[c]{@{}c@{}}  Model  \end{tabular}} & Distance & Ratio  & BLiMP & Latent Rep. & Zero Shot & RPM\\
        & Effect ($R^2$) & Effect ($R^2$)& (Acc.) &  \multicolumn{2}{c:}{\begin{tabular}[c]{@{}c@{}}  (Average Spearman's Correlation)  \end{tabular}}  & (Acc.)\\ 
        \hline
        Amber-7B & 0.913 & 0.591 & 0.794  & 0.083 & 0.250  & 0.654\\ 
        \hdashline
        Falcon-7B & 0.928 & 0.838 & 0.817 & -0.116 & 0.180 & 0.730 \\ 
        \hdashline
        Starling-LM-7B-alpha & 0.522 & 0.187 & 0.827  & -0.003 & 0.258 & 0.730 \\ \hdashline
        Llama-2-7B & 0.670 & 0.614 & 0.818 & -0.065 & 0.238 & 0.752 \\ 
        Llama-2-13B & 0.672 & 0.263 & 0.793 & 0.076 & 0.247 & 0.756\\ 

        Llama-3-8B & 0.886 & 0.631 & 0.735 & 0.049 & 0.112 & 0.796 \\ 
        Llama-3-8B-Instruct & 0.903 & 0.745 & 0.788 & 0.004 & 0.128 & 0.810\\

        \hdashline
        Mistral-7B & 0.641 & 0.233 & 0.829 & -0.025 & 0.245 & 0.756 \\ 
        Mistral-7B-Instruct & 0.637 & 0.543 & 0.834  & 0.033 & 0.255 & 0.674\\ \hdashline
        Qwen-0.5B & 0.833 & 0.553 & 0.785 & 0.072 & 0.282 & 0.684\\ 
        Qwen-1.8B & 0.878 & 0.301 & 0.792 & 0.114 & 0.235 & 0.746\\ 
        Qwen-4B & 0.881 & 0.264 & 0.730 & 0.001 & 0.246 & 0.770\\ 
        Qwen-7B & 0.858 & 0.616 & 0.789 & 0.006 & 0.229 & 0.766 \\ 
        Qwen-14B & 0.783 & 0.507 & 0.792 & -0.140 & 0.249  & 0.776 \\ 
        \hdashline
        Pythia-70M & 0.829 & 0.429 & 0.723 & 0.005 & 0.211 & 0.194\\ 
        Pythia-160M & 0.947 & 0.665 & 0.749 & 0.067 & 0.260 & 0.448\\ 
        Pythia-410M & 0.926 & 0.679 & 0.815  & 0.126 & 0.284 & 0.608\\ 
        Pythia-1B & 0.944 & 0.702 & 0.806 & 0.090 & 0.280 & 0.674\\ 
        Pythia-1.4B & 0.933 & 0.764 & 0.819 & 0.074 & 0.283 & 0.730\\ 
        Pythia-2.8B & 0.961 & 0.723 & 0.827 & 0.221 & 0.273 & 0.760\\ 
        Pythia-6.9B & 0.909 & 0.713 & 0.809  & 0.105 & 0.280 & 0.716\\ 
        Pythia-12B & 0.846 & 0.595 & 0.829 & 0.184 & 0.291 & 0.756\\ \hdashline
        Gemini & NA & NA & NA &  \color{blue}{0.311} \tc{*} & NA & NA \\ \hdashline
        GPT-3.5-Turbo & NA & NA & 0.825 & \color{blue}{0.242} \tc{*} & 0.231  & 0.792 \\
        GPT-4 & NA & NA & 0.849 & \color{blue}{0.559} \tc{*} & 0.428 & 0.822  \\ 
        \hline
    \end{tabular}
    }
     \caption{Performance of Pre-trained Language Models on the tasks. Distance Effect: Averaged $R^2$ values of different LLMs when fitting a linear function on the cosine-similarity vs. distance plot. Ratio Effect: Averaged $R^2$ values of different LLMs when fitting a negative exponential function on the cosine-similarity vs. ratio plot. Note: Each value is averaged across all three input types and all model layers to produce one generalizable score. Latent Rep: Average Spearman's Correlation when using the cosine similarity and latent representation-based approach (Note:  \tc{*} refers to the \color{blue}{prompting approaches }\color{black}{for} select models which are gated by APIs, and not the latent representation-based approach), Zero-Shot: Average Spearman's Correlation when using the zero-shot surprisal values, BLiMP: The Benchmark of Linguistic Minimal Pairs for English, RPM: Raven's Progressive Matrices}
    \label{tab:all_results}
\end{table*}

\textbf{Operationalization:} We determine the model's preferred answer for a problem by comparing the surprisal values of the whole sequence (instruction, question, candidate tuple) for each of the candidate options, i.e. the probability of each completed digit representation of a matrix. For the example given in Figure \ref{fig:conversion_rpm}, this would be checking the probability of this sequence (summation of token probabilities) with the correct answer \textbf{(3, 0.6, 0.8)} to the other candidates. A complete list of the prompts used in this paper is given in Appendix \ref{sec:flu}.

\section{Models under consideration}

We evaluate a wide range of language model families, shown in Table \ref{tab:models}. These models are selected based on the following criteria:

 \emph{Public availability}: Open-source models allow us to perform a thorough analysis by accessing the latent representation and the token probability during generation. We follow \citet{faye_models} while choosing PLMs. Although most models in this study are publicly available and open-source, we use three state-of-art commercial PLMs that are gated behind API calls; GPT-3.5-Turbo (pointing to gpt-3.5-turbo-0613 on the OpenAI platform), GPT-4 (pointing to gpt-4-1106 on the OpenAI platform), and Gemini (also referred to as Gemini-1-Pro at the time of writing). The GPT-$x$ model APIs provide token probabilities of the response, allowing us to calculate surprisal, while Gemini does not.
 
 \emph{Availability of multiple sizes}: The availability of model sizes for the same architecture and training paradigms allows us to evaluate the emergent cognitive abilities of the models. We have multiple sizes available for the LLama-2, Qwen, and the Pythia family of models.

 \emph{Availability of intermediate training checkpoints}: This allows us to evaluate the effects of pre-training on the model outputs. Together, the availability of multiple model sizes and intermediate training checkpoints allow us to best evaluate the developmental alignment of PLMs. Amber and Pythia's family of models have available intermediate training checkpoints. While Amber has 360 intermediate checkpoints, the checkpoints are at 4 Billion tokens each and are not at the required granularity. 

\textbf{Pythia Family of models:} Pythia \cite{biderman2023pythia} is one of the first open-source projects with the goal of scientific and transparent model development. It has 8 model sizes ranging from 70 Million to 12 Billion parameters, with each model trained on 286 Billion tokens. The models in the suite are equivalent (in size) to popular decoder architectures like GPT-Neo-(125M, 1.3B, 2.7B) and OPT-(125M, 350M, 1.3B, 2.7B, 6.7B), but with the added benefits of training on a known de-duplicated corpus \cite{gao2020pile}, using the same training order for each model size, and having 154 intermediate checkpoints to study the learning trajectories of PLMs. Thus, the Pythia suite of models is ideal for studying the cognitive and developmental alignment of PLMs to humans. 

All open-source models are obtained from Huggingface \cite{huggingface}, while the gated models are obtained from their respective platforms through API calls. For each model in the Pythia suite, the following intermediate checkpoints are available: [1, 2, 4, 8, ... 512; 1000, 2000, 3000 ... 143000 (exponential increase in checkpoint number until the 512th checkpoint and subsequent progression of 1000 steps until the last checkpoint)], with each checkpoint representing 2 Million tokens seen. \emph{Overall, we test 1232 intermediate checkpoints of the Pythia suite of models across all the tasks.}

\begin{figure*}[h]
\centering
\includegraphics[trim={1 1 1 1}, width=\textwidth]{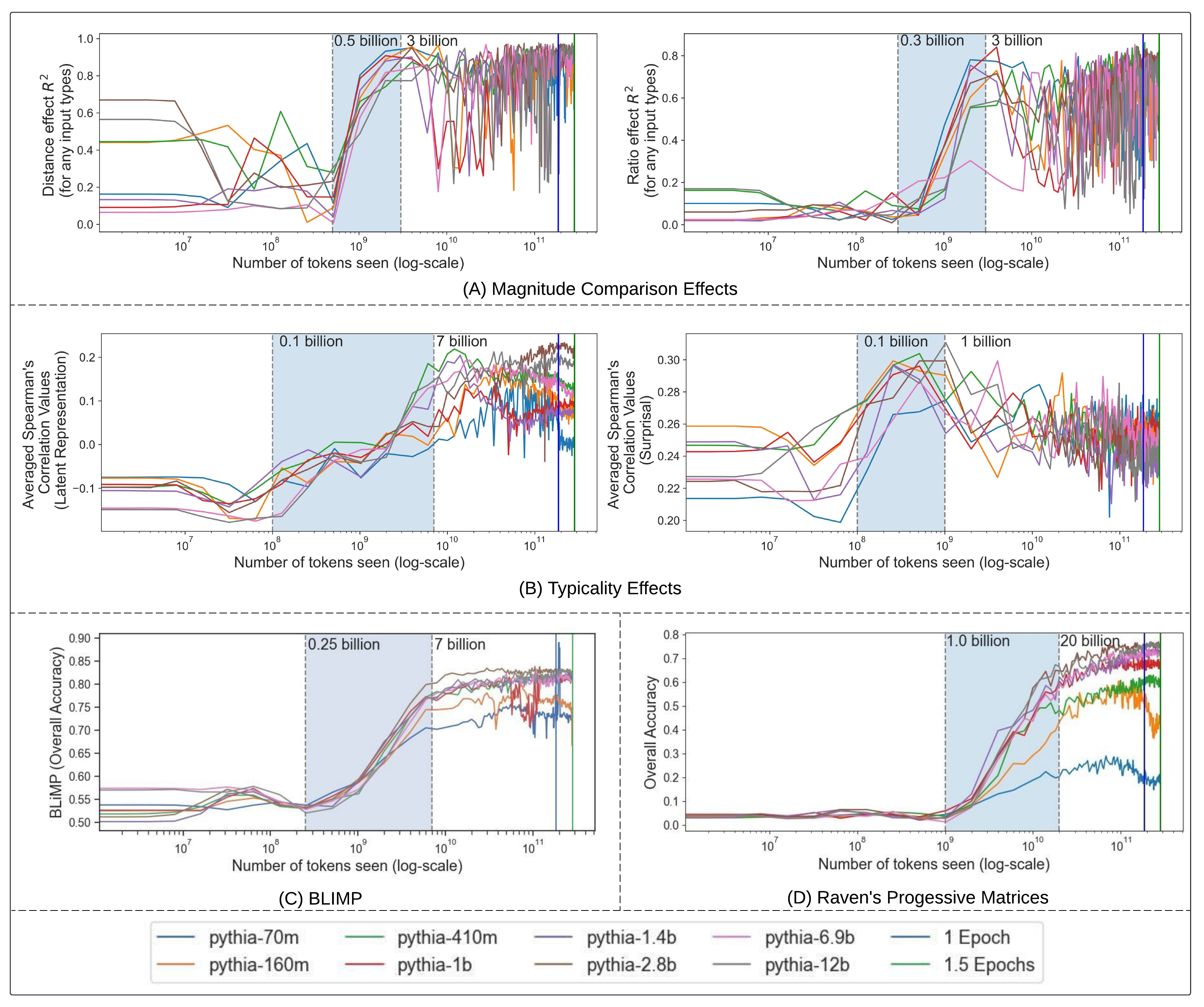}
\caption{Developmental trajectory of the Pythia suite of models on the psychometric intelligence tasks as a function number of tokens seen. We display the x-axis in a log-scaled manner as maximal development occurs in the range of 100 Million to 20 Billion tokens seen for all tasks. The windows of maximal development are illustrated by the blue shading.}
\label{fig:dev_all}
\end{figure*}

\section{Cognitive and developmental alignment of PLMs}
The suite of tasks enables comprehensive evaluation of a variety of PLMs on their cognitive alignment to humans across four domains of psychometric intelligence: numeric abilities, linguistic abilities, concept understanding, and fluid reasoning. Table \ref{tab:all_results} highlights the key results of this evaluation. For the evaluation of conceptual understanding in PLMs, we only report the results for the zero-shot surprisal values and latent representations. This is because we see similar results for zero-shot and few-shot surprisal value-based methods (see comprehensive results in Appendix \ref{sec:typ}).

The \emph{cognitive alignment} of PLMs on psychometrics assessments is summarized below:
\begin{itemize}
    \setlength\itemsep{0em}
    \setlength\parskip{0em}
    \setlength\parsep{0em}
    \item \emph{Numeric abilities}: All PLMs show a human-like distance effect but weakly show a human-like ratio effect. We do not observe any notable changes in alignment with model scaling, indicating the need for the evaluation of future models on this task.
    \item \emph{Linguistic abilities}: The accuracy of the PLMs on the BLiMP linguistic acceptability tasks improves upon increasing the number of parameters. Furthermore, we find that all PLMs are substantially more accurate on morphological tasks over syntactic and semantic tasks (\emph{Accuracy: semantic $<$ syntax $\ll$ morphology}; see Appendix Table \ref{tab:blimp_accuracy}). 
    \item \emph{Concept understanding}: Prompting methods in commercial models perform substantially better than other methods -- closeness judgment and surprisal values -- on all open-source models. In the Pythia suite, we observe that larger models outperform smaller counterparts on the same training data.
    \item \emph{Fluid reasoning}: For all PLM architecture types, larger models outperform their smaller equivalent models.
    \item Despite differences in PLM architecture type, all models of an approximate size of 7 Billion parameters perform comparably.
\end{itemize}

The \emph{developmental alignment} of the PLMs on the tasks is shown in Figure \ref{fig:dev_all}. We make the following key observations:
\begin{itemize}
    \setlength\itemsep{0em}
    \setlength\parskip{0em}
    \setlength\parsep{0em}
    \item \emph{Training endows the ``blank slate'' with requisite structure}: In each assessment, the model ``warm-ups'' in training on a few million/ billion tokens, moving from a ``blank slate'' to possessing the requisite structure. This structure can be thought of as the child's endowment at birth. Development of the four abilities begins only after reaching this state.
    \item \emph{Training shows a region of development}: For all four tasks, we see a window of monotonic development, in which all models gain the respective cognitive abilities. 
    \item \emph{After development, training appears to serve an engineering goal}: After the window of development, the metric becomes unstable once the phenomena are learned. The training appears to serve the engineering goal of loss reduction \cite{Chen2023SuddenDI}.
    This observation is especially pronounced for numeric abilities and conceptual understanding.
    \item \emph{Assessments for Fluid Reasoning and Linguistic Abilities show significant gains with scaling and greater pre-training}: For the assessments of these abilities, we see that the alignment score continues to increase as the PLMs are trained on a greater number of tokens. (Also, morphological performance develops first followed by syntax and then semantics; see Appendix Figure \ref{fig:dev_blimp_all}.) Furthermore, for these abilities, models also show scaling effects, with larger models outperforming smaller ones.
    \item \emph{The relative positions of the windows weakly align with human development}: Variation in the onsets of the windows is weakly consistent with what is known of cognitive development. For example, children acquire language early (i.e., during the preschool years), whereas the onset of improving fluid reasoning is later, when children enter elementary school, and continues for longer, throughout adolescence. Correspondingly, the models significantly develop linguistic abilities while training on 250 Million to 7 Billion tokens, whereas they acquire fluid reasoning abilities later, while training on 1 to 20 Billion tokens.
\end{itemize}

\section{Conclusions}

This paper investigates the appropriateness of using PLMs for human cognitive and developmental modeling. It uses representative assessments of four facets of psychometric intelligence: numeric abilities, linguistic abilities, conceptual understanding, and fluid reasoning. Our experiments show that PLMs develop cognitive abilities purely through their experience in the world, indicating that the cognitive abilities we test are acquirable through mere exposure to language distributions and do not necessarily require innate human-like inductive biases.
Most significantly, we find a window of monotonic development in which all models improve approximately linearly on the four cognitive abilities. Before that window, we interpret training as endowing ``blank slate'' models with the requisite structure for rapid learning. Also notable is the finding of PLM scaling effects for the assessments of linguistic abilities and fluid reasoning. We propose evaluation against these tasks as a prerequisite before treating PLMs as models of human cognition and its development.

\section{Limitations}
Some limitations of the work are as follows: (1) We use an aggregation of psychometric tests for PLMs. The limitations of each test are inherited in the suite of tasks. (2) The alignment scores may be wrongly interpreted when evaluating PLMs with these tasks. Alignment scores show the similarity of PLM outputs to human outputs on psychometric tests and indicate that PLMs do not need explicit neural circuitry for these intelligence tests. We do not suggest these models as proxies for humans in any manner and recommend further testing before use. (3) The developmental alignment of the models points towards the acquisition of human-like performance on the four psychometric assessments in the range of 100 Million to 20 Billion training tokens. This conclusion has two limitations: Pythia is the only suite of models with available intermediate checkpoints and, while unlikely, the observed developmental trajectories might be artifacts of the pre-training order. (4) The psychometric assessments for PLMs are adapted from similar human psychometric tests. Different ways of adaptation may lead to different results. Furthermore, while representative, these assessments are not exhaustive tests of human intelligence. Future work can expand to other tests like spatial and commonsense reasoning. (5) Some open source models like Llama-2 have larger 70 Billion parameter variants but we lack the compute resources to evaluate them. Large open-source models would lead to appropriate comparisons of performance with commercial models like GPT-4. (6) While our work evaluates changes in cognitive alignment with an increase in model size and the number of pre-training tokens, we do not control for different tuning methodologies like instruction tuning and reinforcement learning with human or artificial intelligence feedback. Accounting for different tuning methods is computationally intensive for the 1200+ model checkpoints across 10 architectures. 
\section{Ethical Considerations}
\label{sec:ethical_cons}
All tasks and corresponding datasets have low ethical risks and none expose sensitive information. Additionally, we obtain approval from the authors of each dataset for their use and release. There are no major risks associated with conducting this research beyond those associated with working with PLMs. There may be risks in misinterpreting the alignment scores when evaluating with the tests. The psychometric analysis of this study is one-way: we look for human performance characteristics and behaviors in PLMs. PLMs are experimental technologies and future work using this research should proceed with caution. Assessment of the tasks indicates PLM alignment -- or the lack thereof -- to human cognitive behavior. Indications of higher human alignment do not indicate an absolute proxy for humans. The goal of tasks in this work is a pre-cursor assessment of PLMs on their ability to act as cognitive models. Therefore, researchers and users should perform more tests before use. 

\bibliography{anthology,custom}
\bibliographystyle{acl_natbib}

\appendix
\section{Computational Resources}
The models are evaluated on Nvidia A100 GPUs with 80 GB RAM. The evaluation in this paper cumulatively takes 1600 GPU hours. We use the provided APIs by OpenAI and Google for models of the GPT-X family and Gemini respectively.

\section{Extended set of experiments}
\label{sec:appendix}
\subsection{Numeric abilities: Magnitude comparison effects}
\label{nums_appendix}
\begin{table*}[!h]
  \centering
  
  \resizebox{\textwidth}{!}{%
  \begin{tabular}{lcccccccc}
    \hline
    Model & Distance Effect & Ratio Effect & Size Effect & MDS Stress & MDS Correlation & Range (Sim) & Max (Sim) \\
    \hline
    Amber-7B & 0.913 & 0.591 & 0.607 & 0.157 & 0.572 & 0.008 & 0.995 \\
    \hdashline
    Falcon-7B & 0.928 & 0.838 & 0.725 & 0.183 & 0.655 & 0.286 & 0.779 \\
    \hdashline
    Starling-LM-7B-alpha & 0.522 & 0.187 & 0.494 & 0.320 & 0.305 & 0.001 & 0.995 \\
    \hdashline
     Llama-2-7B & 0.670 & 0.614 & 0.535 & 0.122 & 0.547 & 0.016 & 0.983 \\ 
        Llama-2-13B & 0.672 & 0.263 & 0.421 & 0.234 & 0.372 & 0.002 & 0.999 \\ 

    Llama-3-8B & 0.886 & 0.631 & 0.403 & 0.304 & 0.587 & 0.003 & 0.996 \\ 
        Llama-3-8B-Instruct & 0.903 & 0.745 & 0.409 & 0.324 & 0.512 & 0.005 & 0.995 \\ 
    \hdashline
    Mistral-7B & 0.641 & 0.233 & 0.244 & 0.287 & 0.425 & 0.001 & 0.996 \\ 
        Mistral-7B-Instruct & 0.637 & 0.543 & 0.182 & 0.317 & 0.512 & 0.001 & 0.992 \\

    \hdashline
     Qwen-0.5B & 0.833 & 0.553 & 0.215 & 0.246 & 0.679 & 0.064 & 0.911 \\ 
        Qwen-1.8B & 0.878 & 0.301 & 0.330 & 0.198 & 0.328 & 0.107 & 0.902 \\ 
        Qwen-4B & 0.881 & 0.264 & 0.330 & 0.215 & 0.581 & 0.160 & 0.763 \\ 
        Qwen-7B & 0.858 & 0.616 & 0.257 & 0.153 & 0.636 & 0.129 & 0.734 \\
        Qwen-14B & 0.783 & 0.507 & 0.206 & 0.248 & 0.369 & 0.138 & 0.710 \\ 
    \hdashline
    Pythia-70M & 0.829 & 0.429 & 0.418 & 0.204 & 0.463 & 0.060 & 0.949 \\ 
        Pythia-160M & 0.947 & 0.665 & 0.382 & 0.231 & 0.715 & 0.042 & 0.970 \\ 
        Pythia-410M & 0.926 & 0.679 & 0.393 & 0.210 & 0.710 & 0.041 & 0.972 \\ 
        Pythia-1B & 0.944 & 0.702 & 0.470 & 0.196 & 0.725 & 0.037 & 0.973 \\ 
        Pythia-1.4B & 0.933 & 0.764 & 0.600 & 0.203 & 0.658 & 0.022 & 0.983 \\ 
        Pythia-2.8B & 0.961 & 0.723 & 0.459 & 0.256 & 0.737 & 0.009 & 0.993 \\ 
        Pythia-6.9B & 0.909 & 0.713 & 0.535 & 0.195 & 0.663 & 0.013 & 0.990 \\ 
        Pythia-12B & 0.846 & 0.595 & 0.540 & 0.189 & 0.620 & 0.007 & 0.993 \\ 

    \hline
  \end{tabular}
  }
  \caption{Magnitude Comparison effects. Distance Effect: Averaged $R^2$ values of different LLMs when fitting a linear function on the cosine-similarity vs distance plot. Size Effect: Averaged $R^2$ values of different LLMs when fitting a linear function on the cosine-similarity vs size-difference plot. Ratio Effect: Averaged $R^2$ values of different LLMs when fitting a negative exponential function on the cosine-similarity vs ratio plot. Note: Each value is averaged across all three input types and all model layers to produce one generalizable score. MDS Stress: The stress value is a measure of how well the distances between the points in the multidimensional space represent the dissimilarities of the original data points (lower is better). MDS Correlation: Correlation between the MDS solutions and the expected values of human MNL. Range (Sim): This indicates the range of the cosine-similarities. Max (sim): This indicates the maximum similarity between any two numbers. Range and Max (sim) describe the y-axis.}
  \label{tab:number_all_results}
\end{table*}
Physical quantities in the world are encoded as logarithmically scaled magnitude representations \cite{fechnerElementsPsychophysics1860}. While the distance and the ratio effects are the biggest indicators of the presence of such log-scaled magnitude representations and the numerical precision in humans, other human effects also explain the mental number line. These effects are as follows:
\begin{itemize}
    \item \textbf{Distance effect} (refer to figure \ref{fig:introduction0} (A) top): The greater the distance |x-y| between two numbers (x, y), the faster the comparison in humans \cite{MOYER1967}.
    \item \textbf{Size effect}: Given two comparisons of the same distance (i.e., of the same value for |x - y|), the smaller the numbers, the faster the comparison \cite{parkmanTemporalAspectsDigit1971}. 
     \item \textbf{Ratio effect} (refer to figure \ref{fig:introduction0} (A) bottom): The time taken by humans to compare two numbers (x,y) is a decreasing function of the ratio of the larger number over the smaller number $\frac{max(x,y)}{min(x,y)}$ \cite{halberdaIndividualDifferencesNonverbal2008}.
     \item \textbf{Multidimensional scaling:} Along with the three effects, we investigate the consistency of the latent number representations of PLMs with the human MNL using multidimensional scaling \cite{mds_BorgGroenen2005, mds_Ding2018}. MDS recovers the latent representation from the cosine (dis)similarities between the vector representations of all pairs of numbers (for a given LLM, layer, and number format). This is evaluated by the correlation between the positions of the numbers 1 to 9 in the MDS solution and the expected values (log(1) to log (9)) of the human MNL (refer to the correlation value in table \ref{tab:number_all_results}).
\end{itemize}
\begin{figure}[!h]
\centering
\includegraphics[ trim={1cm 1cm 1cm 1cm}, width=0.5\textwidth]{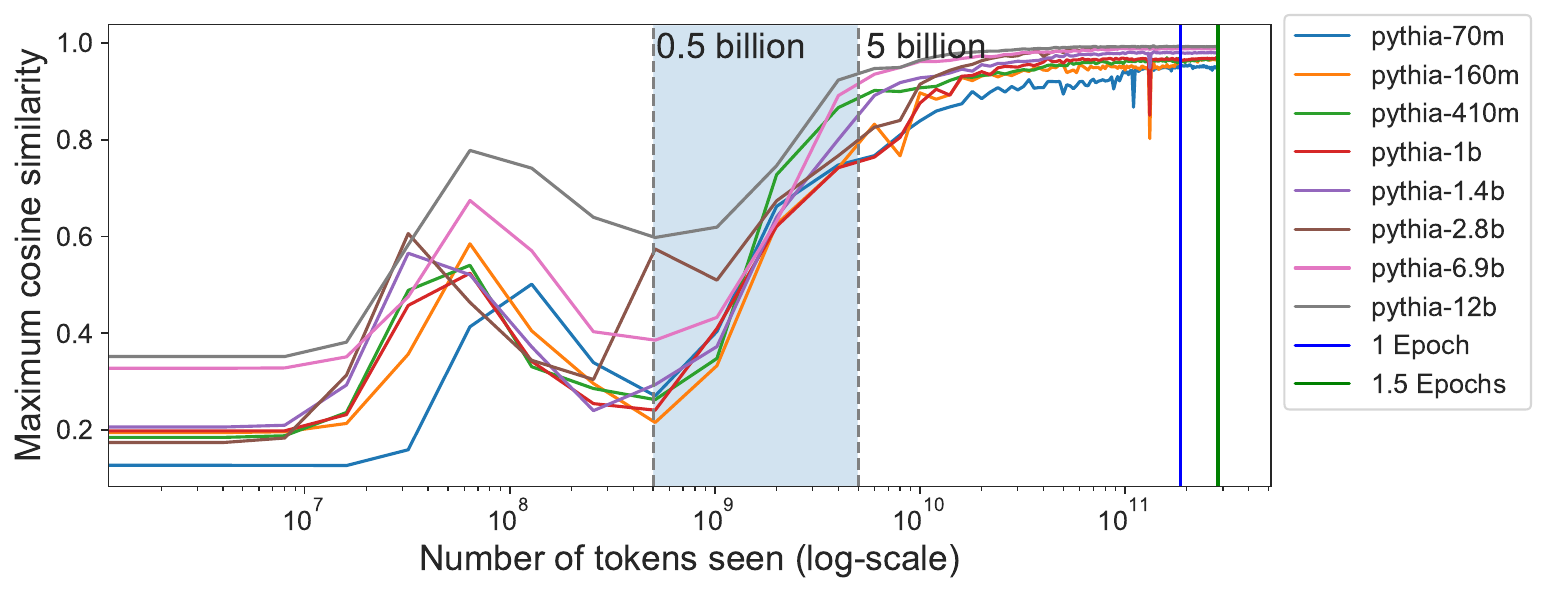}
\caption{Development of the idea of "numbers" in Pythia. The y-axis indicates the maximum cosine similarity between the latent representations of any two number words/ digits.}
\label{fig:max_cosine}
\end{figure}
\begin{figure}[!h]
\centering
\includegraphics[ trim={0cm 1cm 1cm 1cm}, width=0.5\textwidth]{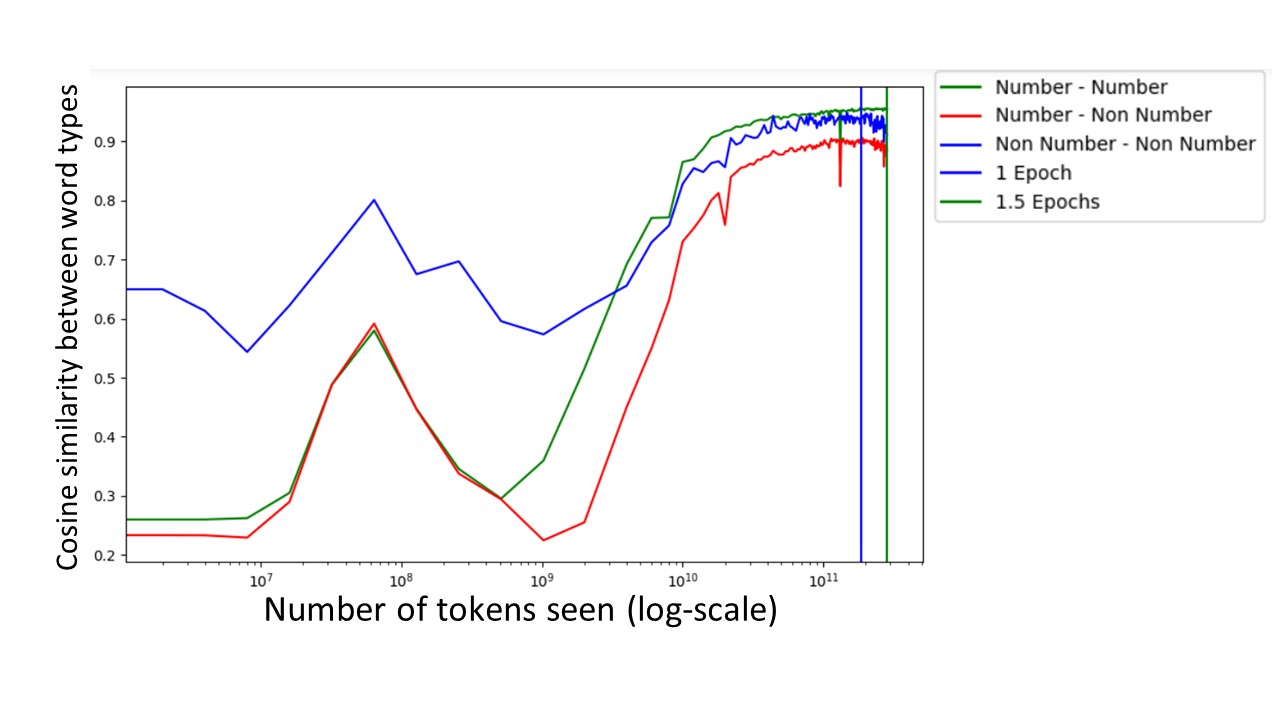}
\caption{Development of the idea of "numbers" in Pythia. The y-axis shows the cosine similarity between word types. The cosine similarity values are averaged over all input types, all model layers, and all model sizes.}
\label{fig:num_num}
\end{figure}

Beyond these effects, we investigate the development of the latent understanding of the concept of "numbers" in the PLMs. As PLMs see more data, the average values of the similarity become larger, indicating that models learn the distinctions among numbers better (refer to figure \ref{fig:max_cosine}). This is further substantiated by figure \ref{fig:num_num}, where the similarities between number words develop to be greater than the similarity between (number, non-number) words and (non-number, non-number) words.

\subsection{Linguistic Abilities}
\label{sec:ling_app}
\begin{table*}[!h]
\centering

\begin{tabular}{lcccccccc}
\hline
\textbf{Model} & \textbf{BLiMP} & \textbf{Syntax} & \textbf{Semantic} & \textbf{Morphology} \\
\hline

Amber-7B & $0.794$ ($\pm$ \small{\textcolor{gray}{0.174}}) & $0.779$ ($\pm$ \small{\textcolor{gray}{0.011}}) & $0.736$ ($\pm$ \small{\textcolor{gray}{0.011}}) & $0.888$ ($\pm$ \small{\textcolor{gray}{0.009}}) \\
\hdashline
Falcon-7B & $0.817$ ($\pm$ \small{\textcolor{gray}{0.173}}) & $0.797$ ($\pm$ \small{\textcolor{gray}{0.011}}) & $0.758$ ($\pm$ \small{\textcolor{gray}{0.011}}) & $0.917$ ($\pm$ \small{\textcolor{gray}{0.008}}) \\
\hdashline
Starling-LM-7B-alpha & $0.827$ ($\pm$ \small{\textcolor{gray}{0.161}}) & $0.799$ ($\pm$ \small{\textcolor{gray}{0.011}}) & $0.788$ ($\pm$ \small{\textcolor{gray}{0.011}}) & $0.938$ ($\pm$ \small{\textcolor{gray}{0.007}}) \\
\hdashline
Llama-2-7B & $0.818$ ($\pm$ \small{\textcolor{gray}{0.165}}) & $0.792$ ($\pm$ \small{\textcolor{gray}{0.011}}) & $0.782$ ($\pm$ \small{\textcolor{gray}{0.011}}) & $0.917$ ($\pm$ \small{\textcolor{gray}{0.008}}) \\
Llama-2-13B & $0.793$ ($\pm$ \small{\textcolor{gray}{0.184}}) & $0.757$ ($\pm$ \small{\textcolor{gray}{0.011}}) & $0.767$ ($\pm$ \small{\textcolor{gray}{0.011}}) & $0.898$ ($\pm$ \small{\textcolor{gray}{0.008}}) \\

Llama-3-8B & $0.735$ ($\pm$ \small{\textcolor{gray}{0.210}}) & $0.708$ ($\pm$ \small{\textcolor{gray}{0.011}}) & $0.651$ ($\pm$ \small{\textcolor{gray}{0.012}}) & $0.871$ ($\pm$ \small{\textcolor{gray}{0.010}}) \\
Llama-3-8B-Instruct & $0.788$ ($\pm$ \small{\textcolor{gray}{0.181}}) & $0.765$ ($\pm$ \small{\textcolor{gray}{0.011}}) & $0.726$ ($\pm$ \small{\textcolor{gray}{0.012}}) & $0.898$ ($\pm$ \small{\textcolor{gray}{0.008}}) \\

\hdashline
Mistral-7B & $0.829$ ($\pm$ \small{\textcolor{gray}{0.174}}) & $0.801$ ($\pm$ \small{\textcolor{gray}{0.011}}) & $0.780$ ($\pm$ \small{\textcolor{gray}{0.010}}) & $0.940$ ($\pm$ \small{\textcolor{gray}{0.007}}) \\
Mistral-7B-Instruct & $0.834$ ($\pm$ \small{\textcolor{gray}{0.149}}) & $0.808$ ($\pm$ \small{\textcolor{gray}{0.011}}) & $0.788$ ($\pm$ \small{\textcolor{gray}{0.011}}) & $0.931$ ($\pm$ \small{\textcolor{gray}{0.008}}) \\

\hdashline
Qwen-0.5B & $0.785$ ($\pm$ \small{\textcolor{gray}{0.176}}) & $0.759$ ($\pm$ \small{\textcolor{gray}{0.012}}) & $0.718$ ($\pm$ \small{\textcolor{gray}{0.012}}) & $0.907$ ($\pm$ \small{\textcolor{gray}{0.008}}) \\
Qwen-1.8B & $0.792$ ($\pm$ \small{\textcolor{gray}{0.162}}) & $0.777$ ($\pm$ \small{\textcolor{gray}{0.012}}) & $0.764$ ($\pm$ \small{\textcolor{gray}{0.011}}) & $0.875$ ($\pm$ \small{\textcolor{gray}{0.010}}) \\
Qwen-4B & $0.730$ ($\pm$ \small{\textcolor{gray}{0.154}}) & $0.694$ ($\pm$ \small{\textcolor{gray}{0.013}}) & $0.728$ ($\pm$ \small{\textcolor{gray}{0.013}}) & $0.814$ ($\pm$ \small{\textcolor{gray}{0.012}}) \\
Qwen-7B & $0.789$ ($\pm$ \small{\textcolor{gray}{0.156}}) & $0.769$ ($\pm$ \small{\textcolor{gray}{0.012}}) & $0.736$ ($\pm$ \small{\textcolor{gray}{0.012}}) & $0.885$ ($\pm$ \small{\textcolor{gray}{0.010}}) \\
Qwen-14B & $0.792$ ($\pm$ \small{\textcolor{gray}{0.144}}) & $0.775$ ($\pm$ \small{\textcolor{gray}{0.012}}) & $0.747$ ($\pm$ \small{\textcolor{gray}{0.012}}) & $0.881$ ($\pm$ \small{\textcolor{gray}{0.010}}) \\

\hdashline
Pythia-70M & $0.723$ ($\pm$ \small{\textcolor{gray}{0.210}}) & $0.701$ ($\pm$ \small{\textcolor{gray}{0.012}}) & $0.628$ ($\pm$ \small{\textcolor{gray}{0.012}}) & $0.872$ ($\pm$ \small{\textcolor{gray}{0.010}}) \\
Pythia-160M & $0.749$ ($\pm$ \small{\textcolor{gray}{0.207}}) & $0.717$ ($\pm$ \small{\textcolor{gray}{0.012}}) & $0.718$ ($\pm$ \small{\textcolor{gray}{0.011}}) & $0.864$ ($\pm$ \small{\textcolor{gray}{0.010}}) \\
Pythia-410M & $0.815$ ($\pm$ \small{\textcolor{gray}{0.169}}) & $0.785$ ($\pm$ \small{\textcolor{gray}{0.011}}) & $0.752$ ($\pm$ \small{\textcolor{gray}{0.011}}) & $0.935$ ($\pm$ \small{\textcolor{gray}{0.007}}) \\
Pythia-1B & $0.806$ ($\pm$ \small{\textcolor{gray}{0.198}}) & $0.782$ ($\pm$ \small{\textcolor{gray}{0.011}}) & $0.728$ ($\pm$ \small{\textcolor{gray}{0.011}}) & $0.935$ ($\pm$ \small{\textcolor{gray}{0.007}}) \\
Pythia-1.4B & $0.819$ ($\pm$ \small{\textcolor{gray}{0.173}}) & $0.792$ ($\pm$ \small{\textcolor{gray}{0.011}}) & $0.768$ ($\pm$ \small{\textcolor{gray}{0.011}}) & $0.931$ ($\pm$ \small{\textcolor{gray}{0.008}}) \\
Pythia-2.8B & $0.827$ ($\pm$ \small{\textcolor{gray}{0.156}}) & $0.800$ ($\pm$ \small{\textcolor{gray}{0.011}}) & $0.782$ ($\pm$ \small{\textcolor{gray}{0.011}}) & $0.925$ ($\pm$ \small{\textcolor{gray}{0.007}}) \\
Pythia-6.9B & $0.809$ ($\pm$ \small{\textcolor{gray}{0.179}}) & $0.792$ ($\pm$ \small{\textcolor{gray}{0.011}}) & $0.750$ ($\pm$ \small{\textcolor{gray}{0.011}}) & $0.913$ ($\pm$ \small{\textcolor{gray}{0.008}}) \\
Pythia-12B & $0.829$ ($\pm$ \small{\textcolor{gray}{0.158}}) & $0.804$ ($\pm$ \small{\textcolor{gray}{0.011}}) & $0.778$ ($\pm$ \small{\textcolor{gray}{0.011}}) & $0.932$ ($\pm$ \small{\textcolor{gray}{0.007}}) \\ \hdashline

Gemini &  NA & NA &  NA & NA \\ \hdashline

GPT-3.5-Turbo & $0.825$ ($\pm$ \small{\textcolor{gray}{0.166}}) & $0.818$ ($\pm$ \small{\textcolor{gray}{0.010}}) & $0.781$ ($\pm$ \small{\textcolor{gray}{0.011}}) & $0.931$ ($\pm$ \small{\textcolor{gray}{0.007}}) \\

GPT-4 & $0.849$ ($\pm$ \small{\textcolor{gray}{0.120}}) & $0.797$ ($\pm$ \small{\textcolor{gray}{0.010}}) & $0.801$ ($\pm$ \small{\textcolor{gray}{0.009}}) & $0.941$ ($\pm$ \small{\textcolor{gray}{0.007}}) \\

\hline
\end{tabular} 
\caption{Accuracy of different language models on the BLiMP linguistic acceptability tasks.}
\label{tab:blimp_accuracy}
\end{table*}

The 12 phenomena tested by BLiMP are as follows:
 \begin{itemize}
     \item Anaphor agreement (morphology): This linguistic phenomenon tests if an anaphor (pronoun) adheres to the antecedent (noun or phrase it refers to) in terms of gender, number, or person. 
     \item Argument Structure (syntax): The argument structure tests the relationship between a verb and its arguments (such as nouns or noun phrases). 
     \item Binding (syntax, semantics): This tests the structural relationship between an anaphor (pronoun) and its antecedent (noun or phrase it refers to). 
     \item Control/ Raising (syntax, semantics): These structures test how semantics differ by syntactical variations of subjects/verbs in subordinate and main clauses.
     \item Determiner-noun agreement (morphology): This tests the agreements of the determiners with the corresponding nouns in number (singular or plural) and sometimes gender (e.g., "his" for masculine nouns, "her" for feminine nouns).
    \item Ellipsis (syntax): This refers to the omission of words from a sentence that can be understood from the context.
    \item Filler-gap (syntax): This tests the syntactic structure of sentences that include phrasal movements (wh-questions, relative clauses).
    \item Irregular forms (morphology): Forms in language that do not follow regular patterns and may need to be memorized. For example, the superlative of good is better, best, and not gooder, goodest.
    \item Island effects (syntax): These test the constraints on syntactic environments where the gap in a filler-gap dependency can occur.
    \item NPI licensing (semantics): This phenomenon tests the constrained situations where negative polarity items like any and ever are limited to the scope of negation.
    \item Quantifiers (semantics): This phenomenon tests the constraints regarding the placement of quantifiers. Specifically, BLiMP looks at superlative quantifiers (such as "at least") that cannot occur within negation, and definite quantifiers and determiners cannot function as subjects in existential "there" constructions.
    \item Subject-verb agreement (morphology): The subject and tense forms of the verb must agree on the number, for example, singular vs plural.
 \end{itemize}

Table \ref{tab:blimp_accuracy} shows that the PLMs are more accurate in morphology than in language syntax and semantics. Most models also perform better on syntactic language features than semantic language features.

\begin{figure*}[h]
\centering
\includegraphics[ width=0.75\textwidth]{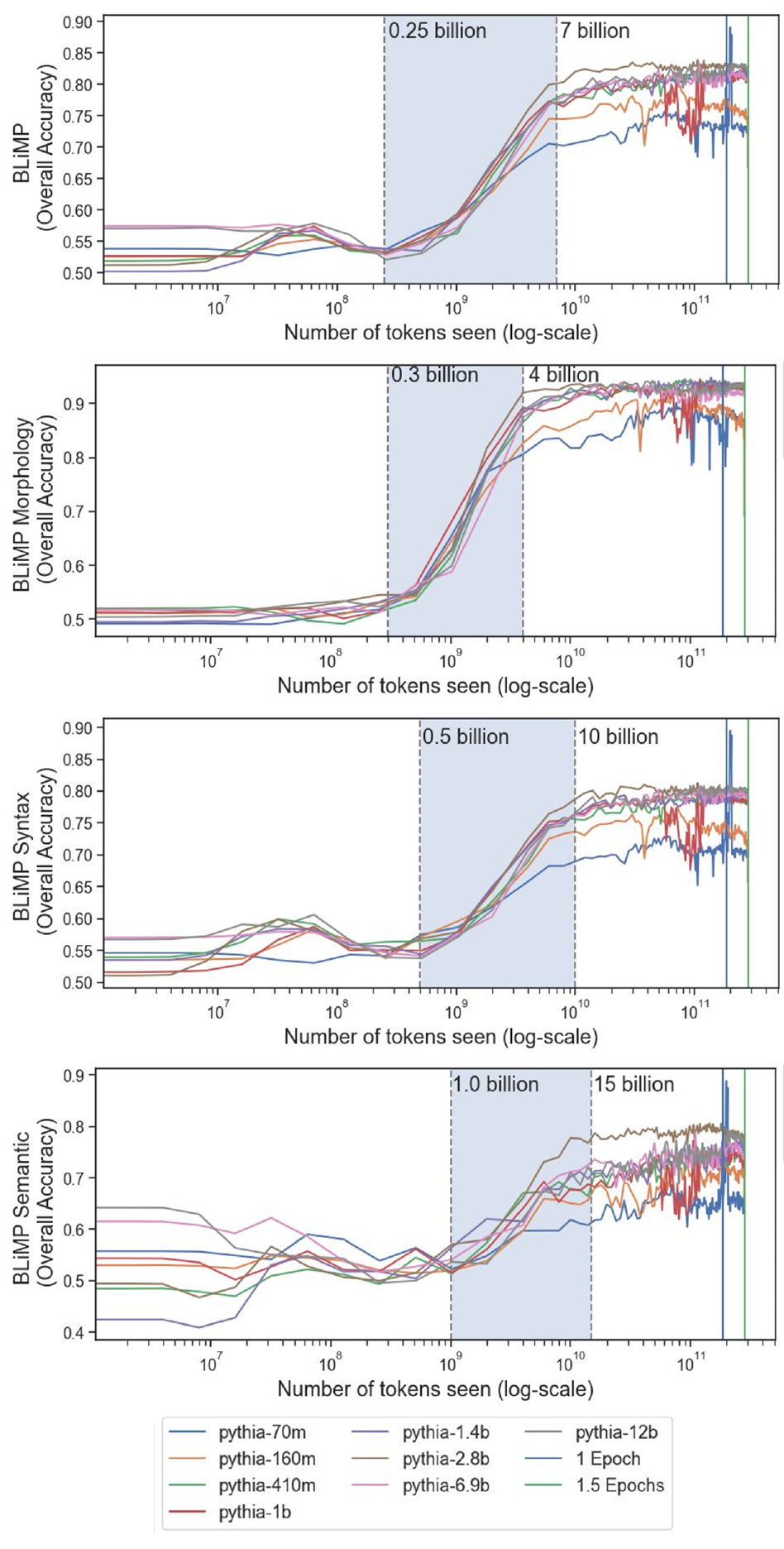}
\caption{Developmental trajectory of the Pythia suite of models on the BLiMP linguistic acceptability tasks.}
\label{fig:dev_blimp_all}
\end{figure*}
\subsection{Conceptual Understanding}
\label{sec:typ}
Table \ref{tab:typicality_all} shows the human alignment of PLMs on their concept understanding for different operationalization methods. We see that Gemini, GPT-3.5-Turbo, and GPT-4 perform better than other models. Furthermore, Surprisal and Prompting-based methods are stronger techniques for evaluating conceptual understanding of models than representation-based methods. Given the higher performance of Prompting methods on three API-based models, we only show the category-wise results for those models. The final prompt design is given in section \ref{sec:appendix_prompt} and table \ref{tab:gpt3_experiments}. Tables \ref{tab:outputs_gemin}, \ref{tab:outputs_gpt}, and \ref{tab:outputs_gpt_4} show Spearman's correlation on the categories along with the standard deviation, the minimum correlation, and the maximum correlation. We perform the same infilling tasks 50 times for each category to account for variations in generations. We note that the models often failed to return all the options in the in-filling task. We discard such situations in our analysis.

Note: Under the closeness judgment protocol, our experiments fail to match up to the performance of the models used by \citet{cog_sci_typicallity}. This is because our choice of open-source models only provides token representations, on which we later perform an aggregation operation. This aggregation operation leads to a loss of information. In contrast,  \citet{cog_sci_typicallity} use sentence-transformer models \cite{sentence-transformers}, which provide singular latent representation for longer text. This variation in experimentation leads to the difference in alignment scores.

\begin{table}[!h]
    \centering
    
    \begin{tabular}{lcccc}
    \hline
      Categories & GPT 3.5 & GPT 4 & Gemini \\ \hline
        bird & 0.183 & 0.536 & 0.353  \\ 
        carpenters tool & 0.418 & 0.679 & 0.610  \\ 
        clothing & 0.022 & 0.594 & 0.155  \\ 
        color & -0.016 & 0.882 & 0.569 \\ 
        dwelling & 0.208 & 0.335 & 0.340 \\ 
        earth formation & 0.251 & 0.496 & 0.155 \\ 
        fabric & 0.48 & 0.708 & 0.504 \\ 
        fish & 0.183 & 0.643 & 0.247 \\ 
        flower & 0.48 & 0.772 & 0.515  \\ 
        flying thing & 0.07 & 0.249 & 0.184  \\ 
        footwear & 0.118 & 0.521 & 0.218  \\ 
        four-legged animal & 0.435 & 0.818 & 0.537  \\ 
        fruit & 0.465 & 0.726 & 0.508  \\ 
        furniture & 0.069 & 0.525 & 0.147  \\ 
        gardeners tool & 0.355 & 0.557 & 0.507 \\ 
        green thing & 0.196 & 0.572 & 0.335 \\ 
        insect & 0.18 & 0.629 & 0.286 \\ 
        instrument & 0.194 & 0.709 & 0.450  \\ 
        kitchen utensil & 0.384 & 0.624 & 0.252  \\ 
        ship & 0.104 & 0.233 & -0.078  \\ 
        snake & 0.177 & 0.419 & 0.328  \\ 
        toy & 0.299 & 0.480 & 0.169  \\ 
        tree & 0.333 & 0.557 & 0.445  \\ 
        vegetable & 0.096 & 0.783 & 0.121  \\ 
        vehicle & 0.17 & 0.381 & 0.033  \\ 
        weapon & 0.348 & 0.421 & 0.239  \\ 
        weather & 0.333 & 0.255 & 0.274  \\ \hline
        Average & 0.242 & 0.559 & 0.311 \\ \hline
    \end{tabular}
    \caption{Typicality effects: Comparing Average Spearman's correlation score across categories from tables \ref{tab:outputs_gemin}, \ref{tab:outputs_gpt}, and \ref{tab:outputs_gpt_4}.}
    \label{tab:outputs_compare}
\end{table}

\begin{table*}[!h]
    \centering
    
    \resizebox{0.7\textwidth}{!}{%
    \begin{tabular}{lc:cccc:c}
    \hline
       Model & Latent  & \multicolumn{4}{c:}{\begin{tabular}[c]{@{}c@{}}  Surprisal  \end{tabular}}  & Prompting \\ 
        &  Representations & \multicolumn{4}{c:}{\begin{tabular}[c]{@{}c@{}}   Values \end{tabular}}  &  \\ 
        &  & Zero-shot & One-shot & Two-shot & Three-shot &  \\ \hline
        Amber-7B & 0.083 & 0.250 & 0.227 & 0.261 & 0.247 & NA \\ \hdashline
        
        Falcon-7B & -0.116 & 0.180 & 0.215 & 0.242 & 0.200 & NA \\ \hdashline
        Starling-LM-7B-alpha & -0.003 & 0.258 & 0.211 & 0.215 & 0.235 & NA \\ \hdashline
        Llama-2-7B & -0.065 & 0.238 & 0.213 & 0.202 & 0.207 & NA \\ 
        Llama-2-13B & 0.076 & 0.247 & 0.163 & 0.183 & 0.170 & NA \\ 

        Llama-3-8B & 0.049 & 0.112 & 0.167 & 0.193 & 0.196 & NA \\ 
        Llama-3-8B-Instruct & 0.004 & 0.128 & 0.213 & 0.261 & 0.262 & NA \\

        \hdashline
        Mistral-7B & -0.025 & 0.245 & 0.219 & 0.261 & 0.257 & NA \\ 
        Mistral-7B-Instruct & 0.033 & 0.255 & 0.192 & 0.204 & 0.235 & NA \\ \hdashline
        Qwen-0.5B & 0.072 & 0.282 & 0.264 & 0.288 & 0.250 & NA \\ 
        Qwen-1.8B & 0.114 & 0.235 & 0.246 & 0.251 & 0.215 & NA \\ 
        Qwen-4B & 0.001 & 0.246 & 0.217 & 0.252 & 0.193 & NA \\ 
        Qwen-7B & 0.006 & 0.229 & 0.203 & 0.220 & 0.220 & NA \\ 
        Qwen-14B & -0.140 & 0.249 & 0.224 & 0.207 & 0.199 & NA \\ \hdashline
        Pythia-70M & 0.005 & 0.211 & 0.266 & 0.291 & 0.285 & NA \\ 
        Pythia-160M & 0.067 & 0.260 & 0.263 & 0.276 & 0.264 & NA \\ 
        Pythia-410M & 0.126 & 0.284 & 0.235 & 0.282 & 0.242 & NA \\ 
        Pythia-1B & 0.090 & 0.280 & 0.309 & 0.287 & 0.264 & NA \\ 
        Pythia-1.4B & 0.074 & 0.283 & 0.249 & 0.267 & 0.235 & NA \\ 
        Pythia-2.8B & 0.221 & 0.273 & 0.286 & 0.267 & 0.236 & NA \\ 
        Pythia-6.9B & 0.105 & 0.280 & 0.264 & 0.250 & 0.220 & NA \\ 
        Pythia-12B & 0.184 & 0.291 & 0.248 & 0.274 & 0.270 & NA \\ \hdashline
        Gemini & NA & NA & NA & NA & NA & 0.311 \\ \hdashline
        GPT-3.5-Turbo & NA & 0.231 & 0.248 & 0.299 & 0.270 & 0.242 \\ 
        GPT-4 & NA & 0.428 & 0.471 & 0.399 & 0.402 & 0.559 \\ \hline
    \end{tabular}
    }
    \caption{Results for the typicality effects using the three methods}
    \label{tab:typicality_all}
\end{table*}

\begin{table*}[!h]
    \centering
    
    \resizebox{0.7\textwidth}{!}{%
    \begin{tabular}{lcccc}
    \hline
        Categories & Average SpearmanR & Minimum Values & Maximum Values & Std Dev \\ \hline
         bird & 0.353 & -0.156 & 0.582 & 0.144 \\ 
        carpenters tool & 0.610 & 0.417 & 0.885 & 0.104 \\ 
        clothing & 0.155 & -0.104 & 0.523 & 0.141 \\ 
        color & 0.569 & -0.147 & 0.916 & 0.260 \\ 
        dwelling & 0.340 & 0.140 & 0.499 & 0.086 \\ 
        earth formation & 0.155 & -0.449 & 0.494 & 0.191 \\ 
        fabric & 0.504 & 0.125 & 0.811 & 0.168 \\ 
        fish & 0.247 & -0.505 & 0.611 & 0.265 \\ 
        flower & 0.515 & -0.183 & 0.779 & 0.208 \\ 
        flying thing & 0.184 & -0.068 & 0.602 & 0.193 \\ 
        footwear & 0.218 & -0.340 & 0.569 & 0.215 \\ 
        four-legged animal & 0.537 & 0.225 & 0.689 & 0.099 \\ 
        fruit & 0.508 & -0.019 & 0.802 & 0.222 \\ 
        furniture & 0.147 & -0.479 & 0.663 & 0.310 \\ 
        gardeners tool & 0.507 & 0.025 & 0.771 & 0.151 \\ 
        green thing & 0.335 & 0.037 & 0.535 & 0.117 \\ 
        insect & 0.286 & -0.121 & 0.635 & 0.193 \\ 
        instrument & 0.450 & 0.092 & 0.832 & 0.175 \\ 
        kitchen utensil & 0.252 & -0.164 & 0.691 & 0.243 \\ 
        ship & -0.078 & -0.414 & 0.277 & 0.179 \\ 
        snake & 0.328 & -0.156 & 0.596 & 0.147 \\ 
        toy & 0.169 & -0.203 & 0.526 & 0.174 \\ 
        tree & 0.445 & 0.257 & 0.585 & 0.073 \\ 
        vegetable & 0.121 & -0.322 & 0.596 & 0.184 \\ 
        vehicle & 0.033 & -0.053 & 0.236 & 0.055 \\ 
        weapon & 0.239 & -0.173 & 0.577 & 0.193 \\ 
        weather & 0.274 & -0.029 & 0.591 & 0.147 \\ \hline
    \end{tabular}
    }
    \caption{Average Spearman's correlation score for each category on 50 runs of each in-filling experiment on the Gemini-Pro model.}
    \label{tab:outputs_gemin}
    
\end{table*}
\begin{table*}[!h]
    \centering
    
    \resizebox{0.7\textwidth}{!}{%
    \begin{tabular}{lcccc}
    \hline
        Categories & Average SpearmanR & Minimum Values & Maximum Values & Std Dev \\ \hline
        bird & 0.183 & -0.209 & 0.552 & 0.209 \\ 
        carpenters tool & 0.418 & -0.162 & 0.858 & 0.282 \\ 
        clothing & 0.022 & -0.321 & 0.540 & 0.192 \\ 
        color & -0.016 & -0.596 & 0.564 & 0.261 \\
        dwelling & 0.208 & -0.053 & 0.400 & 0.123 \\ 
        earth formation & 0.251 & -0.296 & 0.562 & 0.217 \\ 
        fabric & 0.480 & -0.044 & 0.767 & 0.233 \\ 
        fish & 0.183 & -0.326 & 0.690 & 0.280 \\ 
        flower & 0.480 & -0.301 & 0.800 & 0.269 \\ 
        flying thing & 0.070 & -0.181 & 0.377 & 0.149 \\ 
        footwear & 0.118 & -0.439 & 0.581 & 0.241 \\ 
        four-legged animal & 0.435 & -0.264 & 0.869 & 0.292 \\ 
        fruit & 0.465 & -0.006 & 0.868 & 0.241 \\ 
        furniture & 0.069 & -0.325 & 0.447 & 0.195 \\ 
        gardeners tool & 0.355 & -0.311 & 0.796 & 0.294 \\ 
        green thing & 0.196 & -0.337 & 0.572 & 0.211 \\ 
        insect & 0.180 & -0.248 & 0.503 & 0.201 \\ 
        instrument & 0.194 & -0.242 & 0.466 & 0.191 \\ 
        kitchen utensil & 0.384 & -0.610 & 0.797 & 0.334 \\ 
        ship & 0.104 & -0.314 & 0.599 & 0.250 \\ 
        snake & 0.177 & -0.244 & 0.591 & 0.196 \\ 
        toy & 0.299 & -0.210 & 0.603 & 0.180 \\ 
        tree & 0.333 & -0.199 & 0.731 & 0.289 \\ 
        vegetable & 0.096 & -0.191 & 0.542 & 0.172 \\ 
        vehicle & 0.170 & -0.381 & 0.381 & 0.201 \\ 
        weapon & 0.348 & -0.058 & 0.609 & 0.156 \\ 
        weather & 0.333 & -0.425 & 0.662 & 0.236 \\ \hline
    \end{tabular}
    }
    \caption{Average Spearman's correlation score for each category on 50 runs of each in-filling experiment on the GPT-3.5-Turbo model.}
    \label{tab:outputs_gpt}
\end{table*}

\begin{table*}[!h]
    \centering
    
    \resizebox{0.7\textwidth}{!}{%
    \begin{tabular}{lcccc}
    \hline
        Categories & Average SpearmanR & Minimum Values & Maximum Values & Std Dev \\ \hline
        bird & 0.536 & 0.355 & 0.756 & 0.098 \\ 
        carpenters tool & 0.679 & 0.549 & 0.843 & 0.078 \\ 
        clothing & 0.594 & 0.350 & 0.751 & 0.100 \\ 
        color & 0.882 & 0.813 & 0.952 & 0.035 \\ 
        dwelling & 0.335 & 0.183 & 0.497 & 0.070 \\ 
        earth formation & 0.496 & 0.373 & 0.628 & 0.061 \\ 
        fabric & 0.708 & 0.583 & 0.801 & 0.052 \\ 
        fish & 0.643 & -0.237 & 0.817 & 0.218 \\ 
        flower & 0.772 & 0.629 & 0.869 & 0.057 \\ 
        flying thing & 0.249 & -0.118 & 0.704 & 0.221 \\ 
        footwear & 0.521 & 0.191 & 0.721 & 0.112 \\ 
        four-legged animal & 0.818 & 0.634 & 0.906 & 0.056 \\ 
        fruit & 0.726 & 0.567 & 0.868 & 0.069 \\ 
        furniture & 0.525 & 0.381 & 0.605 & 0.055 \\ 
        gardeners tool & 0.557 & 0.314 & 0.757 & 0.098 \\ 
        green thing & 0.572 & 0.444 & 0.709 & 0.050 \\ 
        insect & 0.629 & 0.451 & 0.871 & 0.103 \\ 
        instrument & 0.709 & 0.585 & 0.885 & 0.064 \\ 
        kitchen utensil & 0.624 & 0.358 & 0.750 & 0.075 \\ 
        ship & 0.233 & -0.346 & 0.618 & 0.232 \\ 
        snake & 0.419 & 0.002 & 0.575 & 0.108 \\ 
        toy & 0.480 & 0.277 & 0.675 & 0.111 \\ 
        tree & 0.557 & 0.300 & 0.781 & 0.106 \\ 
        vegetable & 0.783 & 0.413 & 0.892 & 0.102 \\ 
        vehicle & 0.381 & 0.166 & 0.699 & 0.119 \\ 
        weapon & 0.421 & 0.268 & 0.650 & 0.082 \\ 
        weather & 0.255 & 0.122 & 0.357 & 0.061 \\ \hline
    \end{tabular}
    }
    \caption{Average Spearman's correlation score for each category on 50 runs of each in-filling experiment on the GPT-4 model.}
    \label{tab:outputs_gpt_4}
\end{table*}

\begin{table*}[!htb]
\centering
    
\resizebox{\textwidth}{!}{%
    
    \begin{tabular}{lcc}
    \hline
       Prompt region &  Description & Actual prompt\\ 
       \hline
       Guidelines & Describe the overall idea of typicality to the model and the task guidelines & Appendix \ref{sec:appendix_prompt}\\
       Query & This is the actual fill-in-the-blanks task & The \_\_\_ is a "Category-Name"\\
       Options & List of items in a randomized order and separated by a new line & ---
       \\ \hline
    \end{tabular}
    }
    \caption{Prompt design for evaluating typicality effects in models bigger than 30 billion parameters.}
    \label{tab:gpt3_experiments}
\end{table*}

\newpage
\onecolumn
\includepdf[scale=0.8,pages=1,pagecommand={ \subsubsection{Conceptual Understanding - Final Prompt} \label{sec:appendix_prompt} \thispagestyle{empty}}, fitpaper=true]{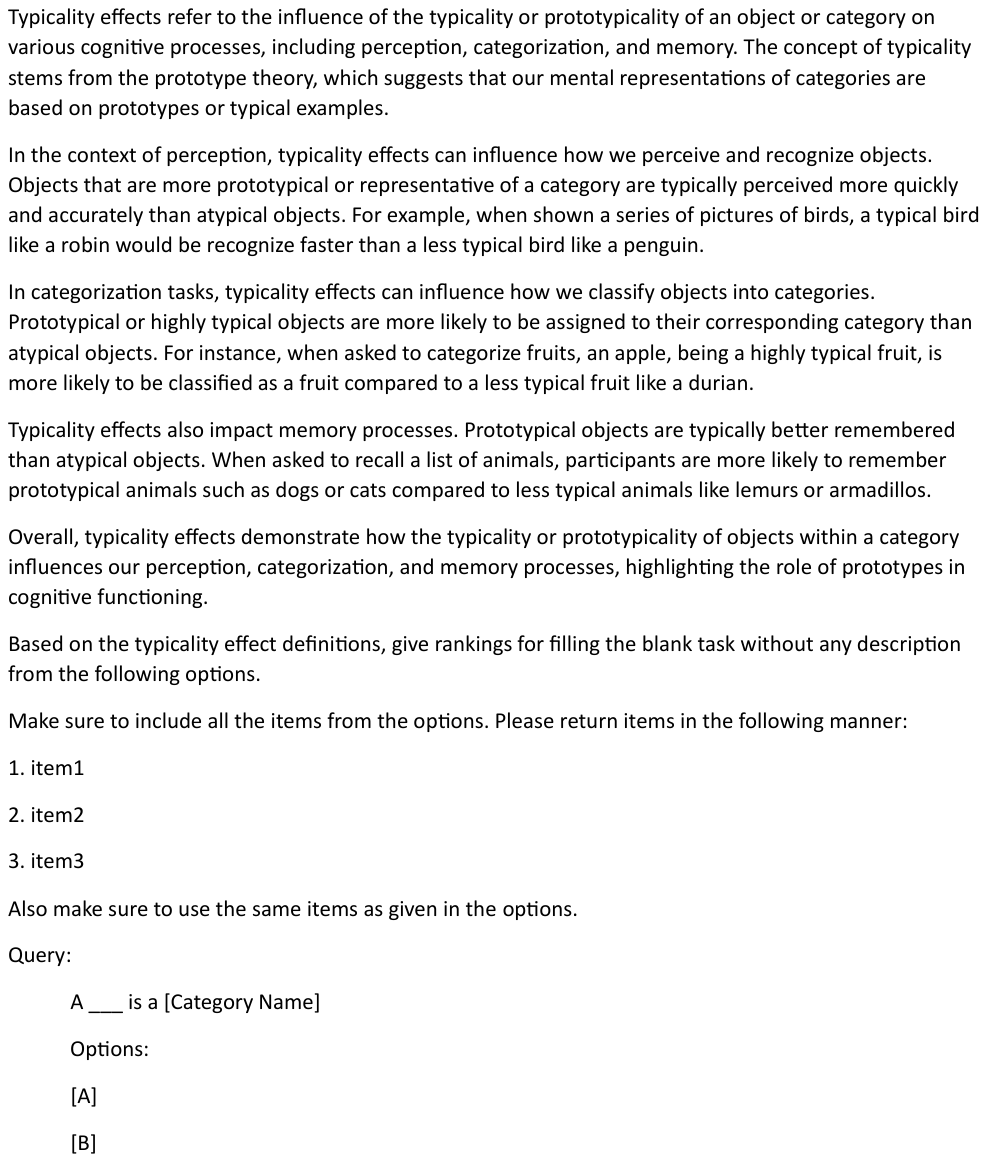}
\newpage

\subsection{Fluid Reasoning}
\label{sec:flu}
Humans cannot completely operate without relying on prior experience. The pervasive role of prior knowledge in shaping cognition is a foundational tenet of the cognitive revolution. However, “Fluid intelligence” is the ability to solve novel and abstract problems \cite{raven2003raven}. It is a core cognitive ability, closely related to other domain-general cognitive abilities like working memory, and executive function, both correlationally \cite{Conway2002ALV} and in terms of the underlying neural correlates (i.e., in the prefrontal cortex) \cite{Burgess2011JournalOE}. It is distinguished from crystallized intelligence, which is composed of the domain-specific knowledge and skills one acquires through one's lifetime \cite{Hartshorne2015WhenDC}. This distinction is a classic one in psychology \cite{carroll1993human}.
\subsubsection{Scholastic Assessment Test analogy questions}
Previous work has shown that fluid reasoning correlates with analogical reasoning \cite{GOSWAMI198673, snow1984topography, cattell1987intelligence}. 
AI, ML, and NLP research has focused on analogical reasoning because this requires many componential abilities: syntactic parsing, semantic understanding, categorization, inductive reasoning, mathematical reasoning, and so on \cite{mil}. Research on the cognitive alignment of PLMs has focused on performance on the 374 Scholastic Assessment Tests (SAT) analogy questions by \citet{turney2005measuring}. Despite being broadly used in literature \cite{turney2005measuring,turney2010frequency,hendrickx2019semeval, webb2023emergent}, our pilot experiments show that PLMs like GPT-3.5-Turbo, GPT-4, and Gemini perform nearly at ceiling on this test, while other open source models perform poorly on the same test. This hints that the set of questions in the test may be part of the GPT-X/ Gemini training or tuning data.

\textbf{Operationalization}: Each problem is of the form A:B::?, with answer choices containing candidates for C:D. We evaluate the performance of models in three ways:
\begin{itemize}
    \setlength\itemsep{0em}
    \setlength\parskip{0em}
    \setlength\parsep{0em}
     \item Closeness judgment problem: Calculate the cosine similarity between the obtained latent representations for the member and the category. This requires models where the latent representations are readily available. These cosine similarities are calculated in different ways:
     \begin{itemize}
         \item 3-cos-add: cos( vector(D),vector(C) - vector(A) + vector(B))
         \item 3-cos-mul: cos(vector(D), vector(B))*cos(vector(D), vector(C))/(cos(vector(D), vector(A))+ e); e is a small constant to prevent overflow.
         \item Concat-cos: cos( [vector(A) || vector(B)] , [vector(C) || vector(D)])
     \end{itemize}
     \item Surprisal values: Calculating the summation of probabilities for each token with the as=to relationship; forming the sequence A is to B as C is to D.
     \item Prompting:  Prompt the models with the following design: Guidelines, Query, and Options. The Guideline highlights the task of solving the analogy problem. The Query consists of A:B. The options are the candidate pairs C:D.
\end{itemize}
\newpage
\includepdf[scale=0.8,pages=1,pagecommand={ \subsubsection{Raven's Progressive Matrices - list of prompts used in experiments} \label{sec:appendix_prompt_rpm} \thispagestyle{empty}}, fitpaper=true]{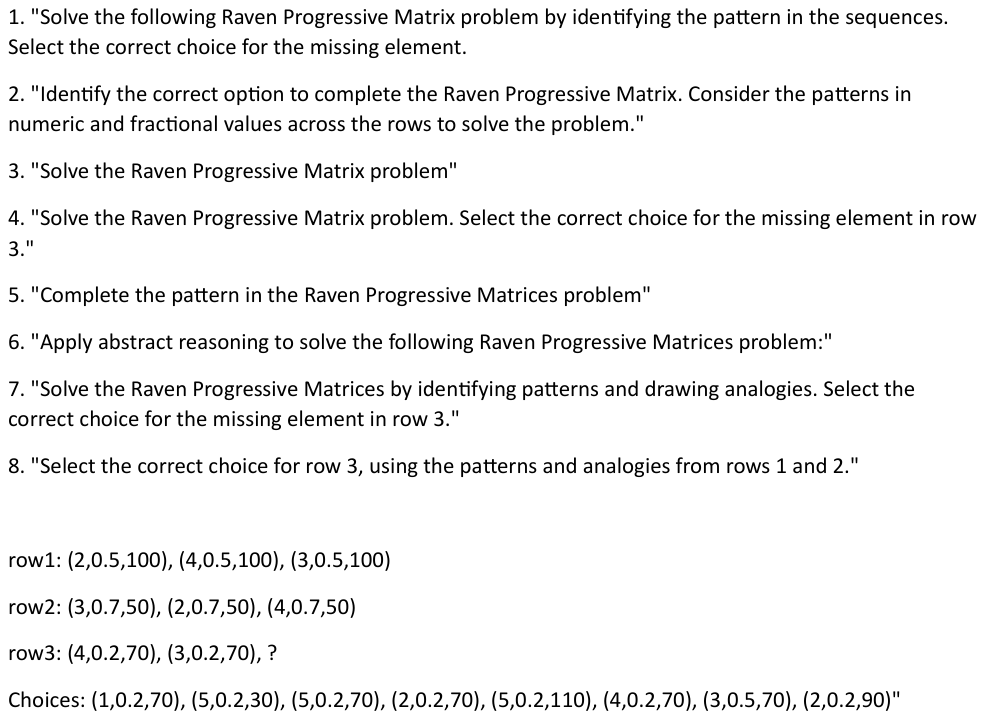}

\end{document}